\documentclass[twoside]{article}

\usepackage[accepted]{aistats2024}
\usepackage{hyperref} 
\hypersetup{colorlinks=true,
            linkcolor = blue,
            urlcolor  = blue,
            citecolor = blue,
            anchorcolor = blue}
\usepackage{url}  
\usepackage{booktabs}  
\usepackage{amsfonts} 
\usepackage{natbib}
\usepackage{xcolor}         % colors
\usepackage{algorithm}
\usepackage{graphicx}
\usepackage{subcaption}
\usepackage{svg}
\usepackage{booktabs} 
\usepackage{algorithmic}
\usepackage{natbib}
\usepackage{lipsum}
\usepackage{amsmath}
\usepackage{siunitx}
\begin{document}



\twocolumn[

\aistatstitle{Active Learning for Abrupt Shifts Change-point Detection via Derivative-Aware Gaussian Processes}

\aistatsauthor{ Hao Zhao\(^{\dagger}\)  \And Rong Pan\(^{\dagger}\)}

\aistatsaddress{ School of Computing and Augmented Intelligence\\
Arizona State University } ]

\begin{abstract}
Change-point detection (CPD) is crucial for identifying abrupt shifts in data, which influence decision-making and efficient resource allocation across various domains. To address the challenges posed by the costly and time-intensive data acquisition in CPD, we introduce the Derivative-Aware Change Detection (DACD) method. It leverages the derivative process of a Gaussian process (GP) for Active Learning (AL), aiming to pinpoint change-point locations effectively. DACD balances the exploitation and exploration of derivative processes through multiple data acquisition functions (AFs). By utilizing GP derivative mean and variance as criteria, DACD sequentially selects the next sampling data point, thus enhancing algorithmic efficiency and ensuring reliable and accurate results. We investigate the effectiveness of DACD method in diverse scenarios and show it outperforms other active learning change-point detection approaches.
\end{abstract}

%%%%%%%%%%%%%%%%%%%%%%%%%%%%%%%%%%%%%%%%%%%%%%%%%%%%%%%%%%%%
\section{INTRODUCTION}

Change-point detection (CPD) deals with identifying abrupt changes or jump discontinuities occurring in a sequence of observed samples from a temporal or spatial stochastic process. CPD is applicable across a wide variety of domains, including physical phenomena \citep{hayashi2019active}, environmental monitoring \citep{verbesselt2010detecting, oelsmann2022bayesian}, process control \citep{ahmadzadeh2018change}, and finance \citep{pepelyshev2015real}, etc. For example, the stock market exhibits a steep change in value when the financial crisis increases volatility, and a manufacturing or production system may experience sudden and irregular changes or fluctuations when the production process begins to fail or when there are changes in system components. The detection of such change-points is critical to the reactions to the underlying regime dynamics since these abrupt changes can result in a detrimental impact on any predictive analytics of the system. CPD methods can be categorized into online detection methods, which detect changes as they occur, and offline detection methods, which retrospectively identify changes after receiving all samples. Here, our work focuses on the offline detection.

Most existing CPD approaches rely on certain parametric assumptions about the statistical model of the data and the characteristics of the changes. Examples of such parametric approaches include generalized likelihood ratio (GLR) \citep{appel1983adaptive}, cumulative sum (CUSUM) tests \citep{basseville1993detection}, and Bayesian change-point analysis \citep{pan2012bayesian, truong2020selective}. However, these parametric models share a common limitation: their performance heavily depends on whether the actual data follows the assumed distribution. If the chosen model does not adequately capture the actual underlying process, the CPD performance may be compromised. Alternatively, non-parametric approaches to CPD predominantly rely on hypothesis testing, declaring a change-point when a test statistic surpasses a defined threshold. And Gaussian process (GP), as a non-parametric model, allows itself to be more robust and adaptable to various data types and change-point scenarios. For a more complete overview of the CPD methods, see \citep{truong2020selective,van2020evaluation,aminikhanghahi2017survey}.

In certain scenarios, data acquisition can be expensive, time-consuming, and resource-intensive. Such scenarios include the identification of phase transitions in material science \citep{schiepek2020convergent}, the assessment of seafloor depth \citep{toodesh2021prediction}, evaluating atmospheric and water pollution \citep{ropkins2021early, shen2015impact}, and activities like oil extraction and mining \citep{grzesiek2021method}, etc. These tasks require intricate monitoring and testing to ensure the accuracy of the collected data. To minimize data acquisition costs, it is preferable to select the conditions or locations of measurements interactively based on previous observations instead of opting for random or exhaustive measurements. The Active Learning (AL) technology can help in selecting the most informative data points, allowing for more efficient data acquisition and reducing the overall cost without compromising the performance of CPD. The most recent work that has addressed Active Learning in Change-point Detection is AL-CPD \citep{de2022semi}, which interacts with a human annotator to acquire labels and then identifies new change-points in order to improve the overall performance. Another approach is Active Change-point Detection (ACPD) \citep{hayashi2019active}, which adaptively identifies the next input in a black-box, expensive-to-evaluate function to detect change-points with minimal evaluations. The ACPD method utilizes an existing parametric regime CPD method to compute change scores and a Bayesian Optimization (BO) method to determine the next input. 

In this paper, we provide a novel CPD method that incorporates the derivatives of GPs into AL. We refer our method as Derivative-Aware Change Detection (DACD), as our method will take samples from a derivative process with the aim of detecting sudden shifts in a stochastic process. We begin by fitting a GP model to the initial set of randomly sampled data points, which can adapt and learn the underlying structure of the process. Intuitively, when the process has a discontinuity or abrupt change, its first-order derivative will be maximum at the change location, while the second-order derivative will reveal the function's curvature or concavity \citep{gijbels2005data}. Along with this reasoning, the derivative process of GP can be utilized to formulate the data acquisition function (AF) of AL, in the hope of exploiting and exploring this derivative process to select the most informative data point for correctly identifying an abrupt change. The empirical results from our synthetic experiments with varying data types and change-point characteristics demonstrate the efficiency of the proposed approach.

The main contributions of this paper are summarized as follows:
\begin{itemize}
\item As a novel approach to offline CPD using the GP, DACD does not rely on any underlying parametric assumption about data distribution, thus it is more flexible for detecting abrupt changes in a variety of stochastic processes.
\item We provide a general solution to the CPD problem, in which AL is used to select informative data points for CPD, thus reducing the cost of sampling.
\item We incorporate the GP derivative process into the AF for AL, which can significantly improve the accuracy and speed of the CPD algorithm.
\end{itemize}

In the rest of the paper, we begin with an introduction of related works of CPD in \S\ref{RW}. The background of CPD and GP with derivative processes are presented in \S\ref{B}. In \S\ref{3}, we introduce the procedure of the DACD framework. In \S\ref{4}, we show representative simulation and real-world experiments; and finally, we conclude the paper in \S\ref{5}.

%%%%%%%%%%%%%%%%%%%%%%%%%%%%%%%%%%%%%%%%%%%%%%%%%%%%%%%%%%%%
\section{RELATED WORK}
\label{RW}

In the rich domain of CPD and data-driven optimization, several foundational methods and concepts are considered relevant in understanding the underpinnings and broader context of DACD. This section aims to elucidate the relationship between DACD and a few key methodologies.

AL \citep{settles2009active} frameworks emphasize that the learner starts learning with a small number of initial labeled samples, selects one or a batch of the most useful samples through a certain query function, and asks the oracle (e.g., a human annotator) for the label, and then uses the new knowledge obtained to train the classifier and proceed to the next round query. The problem of choosing samples to label is often referred to as the design of experiments (DOE) in statistics, where the samples are called experiments and their labels are called measurements \citep{myers2016response}. The sequential experimental design and AL both iteratively come up with 'designs' or 'queries' to explore new, most informative samples based on information gained from the current samples \citep{yue2020active}. 

Consider a situation in which we have a large number of data streams with high sampling frequency and high costs for data collection, and the goal is to monitor data streams with partial observations at each time. This known challenge of AL with drifting streaming data \citep{ye2023online} mirrors the objectives of DACD. Another related work is image segmentation \citep{vezhnevets2012active}. CPD resonates with the operation of segregating a time series into diverse segments, drawing parallels to the task of image segmentation, where an image is divided into different segments based on distinct textures or objects. In streaming image analysis, image data are collected over time and it is of tremendous interest to automatically detect abrupt events, such as security breaches from surveillance videos or extreme weather conditions, such as storms, from climatology. In these applications, the data at each time point is the digital encoding of an image.

DACD is also related to BO, an approach for optimizing expensive black-box functions \citep{shahriari2015taking}. It operates by suggesting optimal next inputs based on prior data, thereby effectively minimizing the resource allocation for data acquisition. Often, GPs are employed as the model, attributed to their ability to handle uncertainty. In BO, the employment of gradient data from GPs to accelerate the optimization process is known as the Knowledge-Gradient method \citep{wu2017bayesian}. Furthermore, the gradient of a GP plays a critical role in enhancing optimization and modeling dynamic systems \citep{padidar2021scaling, solak2002derivative}. In these works, it is assumed that both function values and gradient vectors are observable directly. In contrast, our setup does not assume any direct observation of the gradient, and the Gaussian derivative process is obtained by differentiating the kernel function.  Although BO mainly focuses on function optimization, its principle of using past observations to inform future decisions resonates with the fundamental logic behind DACD, and the utilization of derivatives information to guide modeling also illustrates the connection between the two methodologies.

%%%%%%%%%%%%%%%%%%%%%%%%%%%%%%%%%%%%%%%%%%%%%%%%%%%%%%%%%%%%
\section{PROBLEM FORMULATION}
\label{B}

\subsection{Change-point Detection}

Off-line CPD concerns with the identification of the location in a sequence of data where the statistical properties of the data undergo substantial alterations. This field of study posits that these alterations are typically indicative of an underlying change in the data-generating process. One of the common alterations is an abrupt shift in the process. Suppose there are $N$ observations $\mathcal{D} = (X, Y)=\{(x_i, y_i)\}^N_{i=1}$, where the input $x_i \in \mathcal{X}$ (i.e., locations) and the output $y_i \in \mathcal{R}$ (i.e., system responses) follow an unknown function $f(y_i|x_i)$. In the case of a time series, the objective of the CPD algorithm is to identify the change times $\tau_{k}\in\mathcal{X}$ from a collection of $T$ time series denoted as $\{x_1, \cdots, x_T\}$, where $k\in\{1,\cdots, K\}$, and the integer $K$ corresponds to the total number of change times. Thus, the whole time series consists of $K+1$ segments where the underlying state of each segment is different from its adjacent segments.
%%%%%%%%%%%%%%%%%%%%%%%%%%%%%%%%%%%%%%%%%%%%%%%%%%%%%%%%%%%%

\subsection{Gaussian Process}

A GP is a form of stochastic process, comprised of random variables that are indexed through either time or space and each finite group of these variables follows a joint Gaussian distribution. Specifically, the distribution of system output, $f$, is characterized by a mean function $\mu(x)$ and a covariance function $k(x, x^{\prime})$ as follows 
\[\label{eq:1}\tag{1}
f(x)\sim \mathcal{GP}(\mu(x), k(x, x^{\prime})).
\]
An observation of the system, $y_i$, is presumed to be the sum of system response and a noise term, such as $y_i = f(x_i)+\epsilon_i$, where $\epsilon_1,\cdots,\epsilon_N$ are independent and identically distributed Gaussian noise elements from $N(0,\sigma^2)$.
Moreover, to make a prediction, the joint distribution of observations and function values at test locations, represented by $\boldsymbol{x}^\prime$, can be encapsulated as a GP too. Its mean $\mu(\boldsymbol{x}^\prime)$, and covariance $k(\boldsymbol{x}, \boldsymbol{x}^\prime)$ are given by 
\[\label{eq:2}\tag{2}
\begin{aligned}
   \mu(\boldsymbol{x}^\prime) &=\boldsymbol{k}(\boldsymbol{x}, \boldsymbol{x}^{\prime}) \boldsymbol{C}^{-1} \boldsymbol{y}, \\
   k\left(\boldsymbol{\boldsymbol{x}}, \boldsymbol{x}^\prime\right)  &=k(\boldsymbol{x}^{\prime}, \boldsymbol{x}^{\prime})-\boldsymbol{k}(\boldsymbol{x}, \boldsymbol{x}^{\prime})^{\top}\boldsymbol{C}^{-1} \boldsymbol{k}(\boldsymbol{x}, \boldsymbol{x}^{\prime}),
\end{aligned}
\]
where $\boldsymbol{C}=(\boldsymbol{k}(\boldsymbol{x}, \boldsymbol{x})+\sigma^2\boldsymbol{I})$ and $\boldsymbol{I}$ is an identity matrix. 
% This can be optimized utilizing techniques such as BFGS, which exhibit a complexity of $O(N^3)$ per iteration.
%%%%%%%%%%%%%%%%%%%%%%%%%%%%%%%%%%%%%%%%%%%%%%%%%%%%%%%%%%%%

\subsection{Gaussian Process Derivatives }

Given that mathematical differentiation stands as a linear operation, it follows that the derivative of a GP possesses the characteristics of a GP too \citep{rasmussen2006gaussian, inatsu2020active, mchutchon2015nonlinear}. The joint function values and derivative values follow a multi-output GP with mean and kernel functions:
\[\label{eq:3}\tag{3}
\begin{aligned}
\mu^{\nabla}(\boldsymbol{x})&=\left[\begin{array}{c}
\boldsymbol{\mu}(\boldsymbol{x}) \\
\partial_{\boldsymbol{x}} \boldsymbol{\mu}(\boldsymbol{x})
\end{array}\right], \\
k^{\nabla}\left(\boldsymbol{x}, \boldsymbol{x}^{\prime}\right)&=\left[\begin{array}{cc}
\boldsymbol{k}\left(\boldsymbol{x}, \boldsymbol{x}^{\prime}\right) & \left(\partial_{\boldsymbol{x}^{\prime}} \boldsymbol{k}\left(\boldsymbol{x}, \boldsymbol{x}^{\prime}\right)\right)^T \\
\partial_{\boldsymbol{x}}\boldsymbol{k}\left(\boldsymbol{x}, \boldsymbol{x}^{\prime}\right) & \partial^2 k\left(\boldsymbol{x}, \boldsymbol{x}^{\prime}\right)
\end{array}\right],
\end{aligned}
\]
where $\partial_{\boldsymbol{x}}\boldsymbol{k}\left(\boldsymbol{x}, \boldsymbol{x}^{\prime}\right)$ represents the column vector of first-order partial derivatives in $\boldsymbol{x}$, and $\partial^2 k\left(\boldsymbol{x}, \boldsymbol{x}^{\prime}\right)$ is the matrix of mixed partial derivatives in $\boldsymbol{x}$ and $\boldsymbol{x}^{\prime}$.
Then, given the observations $\mathcal{D} =\{(x_i, y_i)\}^N_{i=1}$, the posterior distribution of derivative of $f$ is also a GP, and its mean and covariance (assuming the kernel is twice-differentiable) are respectively given by
\[\label{eq:4}\tag{4}
\begin{aligned}
   \mu^\nabla(\boldsymbol{x}^\prime) = &\partial_{\boldsymbol{x}}\boldsymbol{k}\left(\boldsymbol{x}, \boldsymbol{x}^{\prime}\right)\boldsymbol{C}^{-1} \boldsymbol{y}, \\
   k^\nabla\left(\boldsymbol{x}, \boldsymbol{x}^\prime\right)  = &\partial^2 k\left(\boldsymbol{x}, \boldsymbol{x}^{\prime}\right)-\\
&\left(\partial_{\boldsymbol{x}^{\prime}} \boldsymbol{k}\left(\boldsymbol{x}, \boldsymbol{x}^{\prime}\right)\right)^T\boldsymbol{C}^{-1} \partial_{\boldsymbol{x}}\boldsymbol{k}\left(\boldsymbol{x}, \boldsymbol{x}^{\prime}\right).
\end{aligned}
\]

There is a computational expense associated with GP derivatives, as both their training and inference mechanisms scale at a cubic rate with respect to the number of observations $N$ and dimensionality $D$, expressed as $O(N^3D^3)$.

In this paper, we consider the Radial Basis Function (RBF) kernel as a covariance function, i.e.,
\[\label{eq:5}\tag{5}k_{\mathrm{RBF}}(x, y)=s^2 \exp \left(-\frac{\|x-x^\prime\|^2}{2 \ell^2}\right).\]
Then, the first and mixed partial derivatives of the RBF kernel \citep{johnson2020kernel} are given by
\[\label{eq:6}\tag{6}
\begin{aligned} \frac{\partial k_{\mathrm{RBF}}\left(x, x^\prime\right)}{\partial x_p^{\prime}}=&\frac{x_p-x_p^{\prime}}{\ell^2} k_{\mathrm{RBF}}\left(x, x^{\prime}\right), \\ 
\frac{\partial^2 k_{\mathrm{RBF}}\left(x, x^{\prime}\right)}{\partial x_i \partial x_j^{\prime}}=&\frac{1}{\ell^4}\left(\ell^2 \delta_{i j}-\left(x_i-x_i^{\prime}\right)\left(x_j-x_j^{\prime}\right)\right)\\ &k_{\mathrm{RBF}}\left(x, x^{\prime}\right) .\end{aligned}
\]
We obtained the optimal kernel hyperparameters by applying the L-BFGS optimizer to minimize the negative log marginal likelihoods.

%%%%%%%%%%%%%%%%%%%%%%%%%%%%%%%%%%%%%%%%%%%%%%%%%%%%%%%%%%%%
\section{PROPOSED METHOD}
\label{3}
\subsection{Derivative-Aware Change Detection}

Our methodology is designed for the circumstances where data acquisition incurs a significant cost, thus necessitating the prudent utilization of a limited dataset at the outset. GP emerges as our modeling choice, attributable to their robustness, extensive support, and capacity to act as universal approximators. This universality ensures our method's applicability across various data types, including those exhibiting complex or non-stationary properties. 

After fitting the GP, we proceed to determine the subsequent input location using BO. Within this domain, we employ an AF, $a: \mathcal{X}\rightarrow \mathbb{R}$, which quantifies the desirability of evaluating a particular input location, thereby guiding the selection of the next system sampling point, denoted as $x^{\mathrm{next}}$. Through a GP and an AF $a(x)$, we determine the next input $x^{\mathrm{next}}$ that maximizes the AF:
\[\label{eq:7}\tag{7}
x^{\mathrm{next}}=\underset{x \in \mathcal{X}}{\arg \max }\; a(x).
\]
There are three potential choices for the AF. The first one is the Probability of Improvement (PI) \citep{kushner1964new}, given the currently best observed derivative $f^\nabla(x^+)$, PI evaluates $f^\nabla(x)$ at the point most likely to surpass this value. Consequently, AF is defined as
\[\label{eq:8}\tag{8}
a_{\mathrm{PI}}(x) = \Psi\left(\gamma(x)\right),
\]
where $\gamma(x)=\frac{\mu^\nabla(x)-f^\nabla(x^+)-\xi}{\sigma^\nabla(x)}$,  $\mu^\nabla(x)$ and $\sigma^\nabla(x)= \sqrt{k^\nabla\left(x,x\right)}$ represent the mean and variance of the GP derivative. The hyperparameter $\xi$ modulates the degree of exploration, and $\Psi$ is the cumulative distribution function of a standard Gaussian distribution.
%%%%%%%%%%%%%%%%%%%%%%%%%%%%%%%%%%%%%%%%%
\begin{algorithm}[t]
\caption{DACD}
\label{alg:the_alg}
\renewcommand{\algorithmicrequire}{\textbf{Input:}}
\renewcommand{\algorithmicensure}{\textbf{Output:}}

\begin{algorithmic}[1] 
\REQUIRE 
     \hspace*{\algorithmicindent}{}\\
     Initial set of observations: $\mathcal{D} = \{(x_i, y_i)\}^N_{i=1}$\\
     % Fitted GP: $\mathcal{GP}(\mu_t; k_t(x; x^{\prime}))$\\
     AF: $a$\\ 
     Query budget: $B$\\
\ENSURE Estimated change-point $x_{\hat{\tau}_k}$
\FOR{$i = 1, \cdots, B$}
    \STATE Update the GP: $f(x)|\mathcal{D} \sim\mathcal{GP}(\mu_t, k_t(x, x^{\prime}))$
    \STATE Compute GP derivatives (Eq. (\ref{eq:4}))
    \STATE Select the next sample point $x^{\mathrm{next}}$ based on AF $a$ (Eq. (\ref{eq:8})(\ref{eq:9})(\ref{eq:10}))
    \STATE Query $f$ at $x^{\mathrm{next}}$ to obtain $y^{\mathrm{next}}$ 
    \STATE Append $x^{next}$ to the observations set $\mathcal{D} \leftarrow \mathcal{D} \cup \{(x^{\mathrm{next}}, y^{\mathrm{next}})\}$
    % \IF{Minimum Improvement Threshold achieved}
    %     \STATE Break
    % \ENDIF
\ENDFOR
\STATE Estimate change-points $x_{\hat{\tau}_k}$
\RETURN $x_{\hat{\tau}_k}$
\end{algorithmic}
\end{algorithm}
%%%%%%%%%%%%%%%%%%%%%%%%%%%%%%%%%%%%%%%%%

The second choice is Expected Improvement (EI) \citep{mockus1998application}, which is defined as
\[\label{eq:9}\tag{9}
a_{\mathrm{EI}}(x) = \sigma^\nabla(x)(\gamma(x)\Psi(\gamma(x))+ \Phi(\gamma(x))),
\]
where $\Phi$ is the density function of a standard Gaussian distribution. This policy suggests querying a point where the highest improvement score over the current best function value is expected. 

Finally, we can also use Upper Confidence Bound (UCB) algorithm \citep{srinivas2009gaussian},
\[\label{eq:10}\tag{10}
a_{\mathrm{UCB}}(x) = \mu^\nabla(x) +\lambda\sigma^\nabla(x),
\]
where $\lambda $ is a hyperparameter that balances exploration and exploitation. 

The incorporation of GP derivatives into an AL framework facilitates a balanced interplay between exploration, the selection of data points from areas of high uncertainty, and exploitation, refining the model's comprehension of change-points based on the most informative data. This procedure is iteratively performed until the allocated budget is exhausted, and then output the estimated change location $\hat{\tau}_k$ using a change-point estimation method. The proposed DACD algorithm is presented in {\bf Algorithm \ref{alg:the_alg}}. This algorithm is designed to be able to effectively identify change-points even with a limited number of observations. 

As AL proceeds and an increasing number of samples are collected, the uncertainty of GP variance gradually diminishes. In such a context, taking the EI algorithm as an example, the difference between $\mu^\nabla(x)$ and the best estimated function value $f^\nabla(x^+)$ becomes progressively marginal, which results in a decreasing expected improvement. Given a sufficient quantity of samples, the expected improvement will descend beneath the threshold, signaling the convergence of the algorithm \citep{nguyen2017regret}. Thus, DACD permits us to find the existence of a point $x_{\hat{\tau}_k}$ which corresponds to the actual change-point.

\subsection{Estimation and Evaluation}
The concept of an abrupt shift refers to a significant transformation in the underlying structure of the datasets. To illuminate these hidden changes, we employed the Filtered Derivative method \citep{bertrand2011off}, an analytical tool specifically designed to uncover such shifts. The methodology behind the Filtered Derivative method is based on comparing the empirical means calculated on two distinct sliding windows, which are positioned respectively to the right and left of index $k$, each with a size of $A$. The difference between these two empirical means is associated with a sequence, denoted as $(D(k, A))_{A\leq k\leq n-A}$, and it is defined as follows:
\[\label{eq:11}\tag{11}
D(k, A)=\hat{\mu}(k, A)-\hat{\mu}(k-A, A),
\]
where $\hat{\mu}(k, A)=\frac{1}{A} \sum_{j=k+1}^{k+A} X_j$ represents the empirical mean of $X$, calculated within the sliding window $[k + 1, k + A]$. And we obtain the estimated change-point $\hat{\tau}_k =\arg \max _{k \in[A, n-A]}|D(k, A)|$. The Filtered Derivative approach operates with the underlying presumption that the minimal distance between two consecutive change-points is greater than $2A$, thus, we need some priori knowledge of the datasets to choose the size of $A$.

Given the number of change-points $K$, the method can be further extended to multiple-change-point (MCP) detection, as shown in {\bf Algorithm~\ref{alg:DACD_MCP}}. For each iteration $i$ from $1$ to $K$, the algorithm identifies a change location $\hat{\tau}_k$, suppresses the surrounding region by setting $D(k, A)=0$ for $k$ within an interval around $\hat{\tau}_k$, and subsequently locating the next point that maximizes the absolute difference $|D(k, A)|$. The procedure continually obtains the detected change-points through $K$ iterations, thereby constructing a set $\mathcal{C}=\{x_{\hat{\tau}_1}, x_{\hat{\tau}_2}, \cdots,x_{\hat{\tau}_K} \}$. 

%%%%%%%%%%%%%%%%%%%%%%%%%%%%%%%%%%%%

\begin{algorithm}[t]
\caption{DACD for MCP}
\label{alg:DACD_MCP}
\renewcommand{\algorithmicrequire}{\textbf{Input:}}
\renewcommand{\algorithmicensure}{\textbf{Output:}}

\begin{algorithmic}[1]
    \REQUIRE      \hspace*{\algorithmicindent}{}\\
    Set of sample points from AL loop: \( \mathcal{D} \)\\ 
            Number of change-points: $K$\\
    \ENSURE Change-points set: $\mathcal{C}$    
    \FOR{$i = 1, \cdots, K$}
        \STATE $\hat{\tau}_k= \arg\max_{k \in [A, n-A]} |D(k, A)|$
        \STATE $D(k, A) = 0$ for all $k \in  (\hat{\tau}_k - A, \hat{\tau}_k + A)$
        \STATE Add $x_{\hat{\tau}_k}$ to $\mathcal{C}$
    \ENDFOR
    
    % \STATE Sort $\mathcal{C}$ in ascenting order
    
    \RETURN $\mathcal{C}$
\end{algorithmic}
\end{algorithm}

%%%%%%%%%%%%%%%%%%%%%%%%%%%%%%%%%%%%%

We evaluate the effectiveness of our algorithm by F1 score. 
% \citep{van2020evaluation}, the corresponding equations are: 
% \[\text{Precision}= \frac{\text{TP}}{\text{TP}+\text{FP}}\quad\text{Recall}= \frac{\text{TP}}{\text{TP}+\text{FN}}\]
% \[\text{F1}=\frac{2 \times \text{Precision}\times \text{Recall}}{ \text{Precision}+ \text{Recall}}\]
The true Positives (TP) refer to the count of predicted change-points that fall within a predetermined distance $s$ from true change-points. Predicted change-points that fall outside the distance $s$ from the true change-points are considered False Positives (FP). False Negatives (FN) are those actual change-points that lie beyond the distance $s$ from any predicted change-point. Therefore, precision is understood as the ratio of accurately identified change-points to the total number of detected change-points, while recall signifies the ratio of accurately identified change-points to the overall number of actual change-points.

% In our model, we adopt a distance $s = A/2$, corresponding to the prediction of the change-point within a sliding window. 

%%%%%%%%%%%%%%%%%%%%%%%%%%%%%%%%%%%%%%%%%%%%%%%%%%%%%%%%%%%%

\section{EXPERIMENTS}
\label{4}
%%%%%%%%%%%%%%%%%%%%%%%%%%%%%%%%%%%%%%%%%
\begin{figure*}[t]
    % \vspace{.3in}
  \centering
  \begin{subfigure}[b]{0.99\linewidth}
    \includegraphics[width=\linewidth]{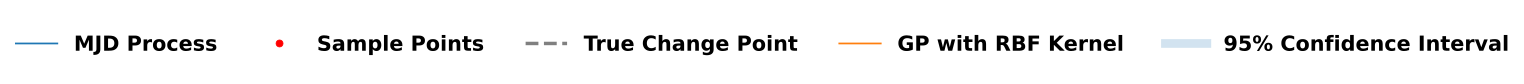}
  \end{subfigure}
  \begin{subfigure}[b]{0.24\linewidth}
    \includegraphics[width=\linewidth]{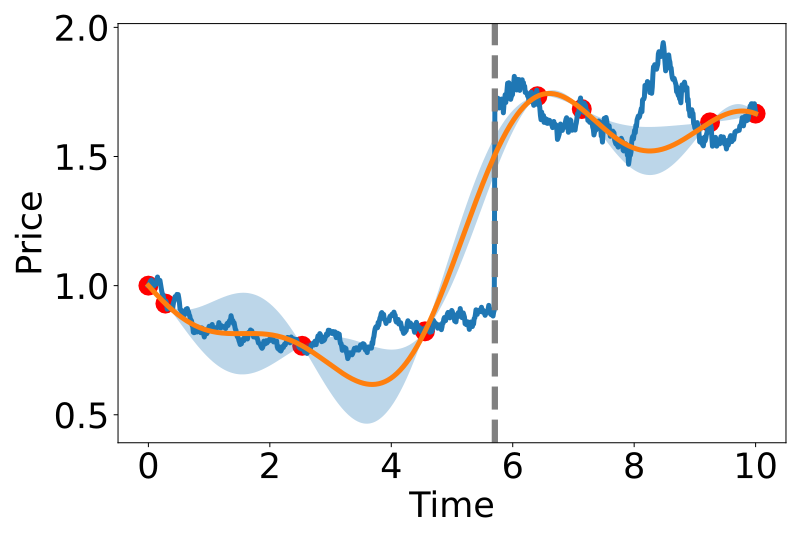}
     \caption{$\text{MJD}_{t_{no}}$}
  \end{subfigure}
  \begin{subfigure}[b]{0.24\linewidth}
    \includegraphics[width=\linewidth]{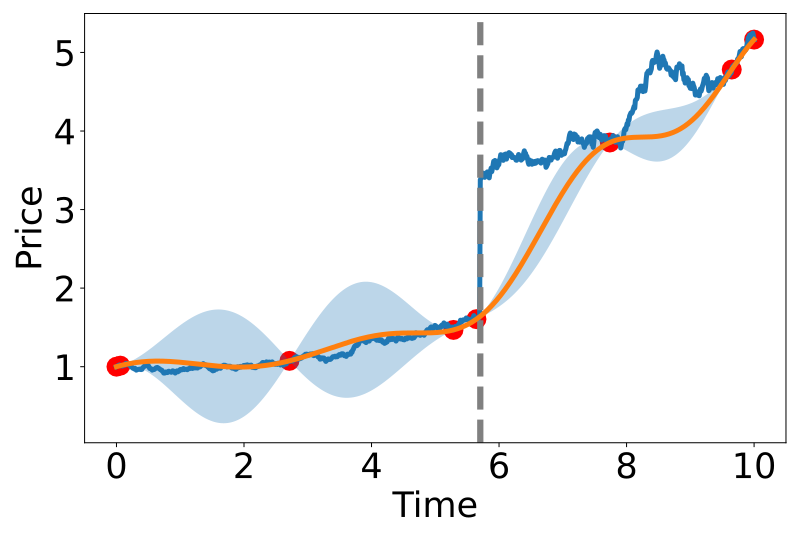}
    \caption{$\text{MJD}_{t_{up}}$}
  \end{subfigure}
  \begin{subfigure}[b]{0.24\linewidth}
    \includegraphics[width=\linewidth]{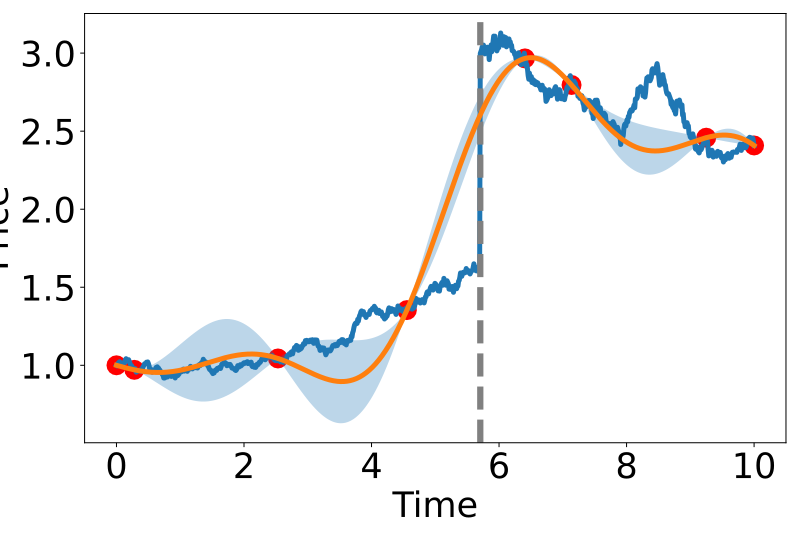}
    \caption{$\text{MJD}_{t_{inv}}$}
  \end{subfigure}
  \begin{subfigure}[b]{0.24\linewidth}
    \includegraphics[width=\linewidth]{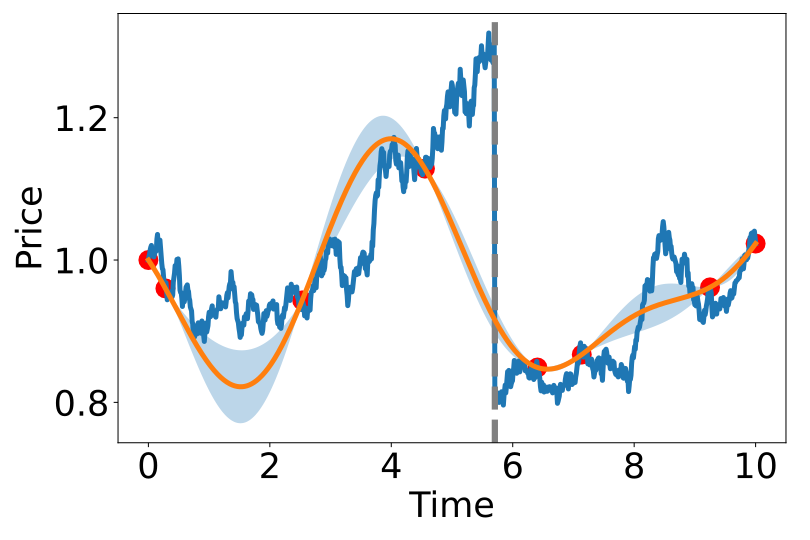}
    \caption{$\text{MJD}_{t_{down}}$}
  \end{subfigure}
    \begin{subfigure}[b]{0.24\linewidth}
    \includegraphics[width=\linewidth]{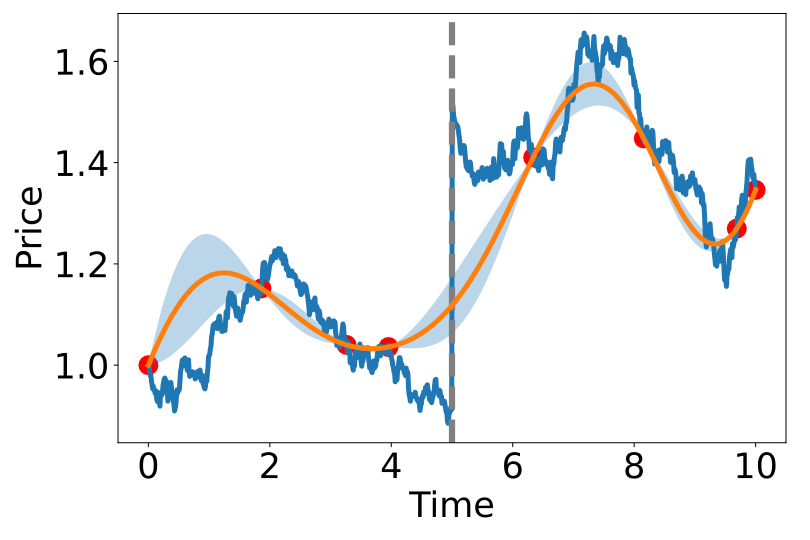}
    \caption{$\text{MJD}_{p_{no}}$}
  \end{subfigure}
  \begin{subfigure}[b]{0.24\linewidth}
    \includegraphics[width=\linewidth]{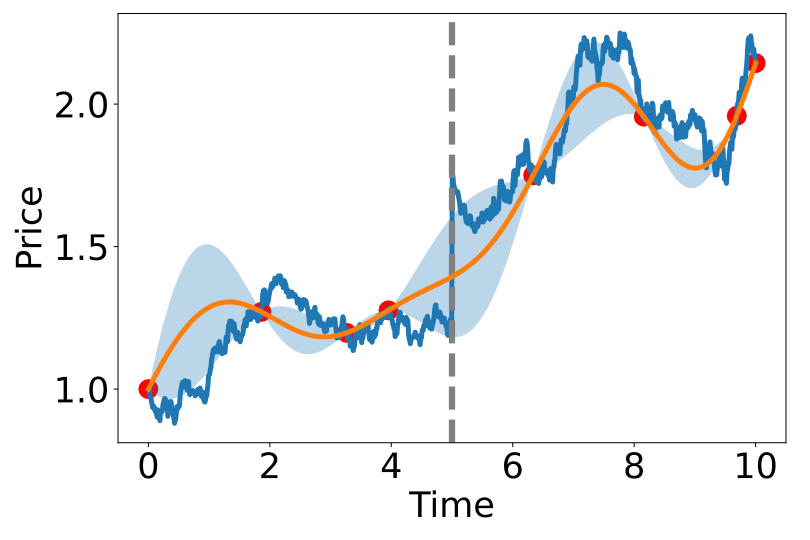}
    \caption{$\text{MJD}_{p_{up}}$ }
  \end{subfigure}
    \begin{subfigure}[b]{0.24\linewidth}
  \includegraphics[width=\linewidth]{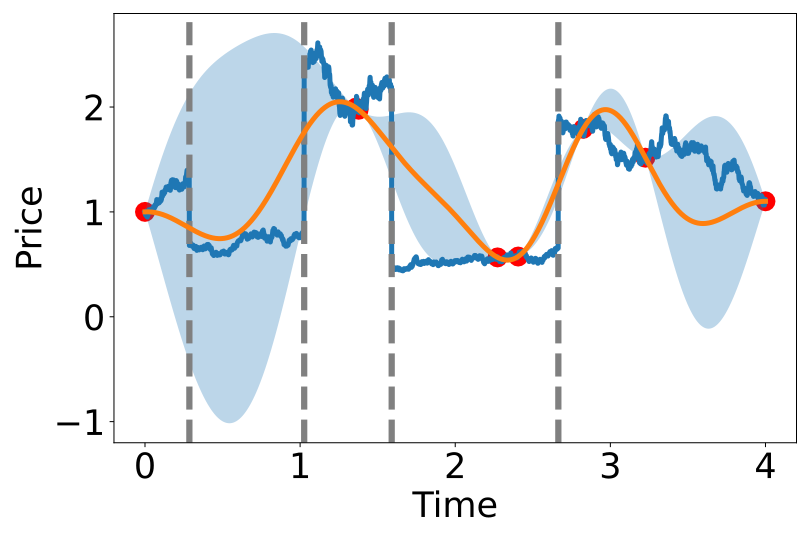}
    \caption{MCP }
  \end{subfigure}
  % \vspace{.3in}
  \caption{Implementation of MJD models and GP regression with RBF kernel after initial 8-point sampling. Blue lines depict MJD processes, red dots indicate sample points, and a grey dashed line marks the ground truth change-point $\tau = 5.7057$ for (a)-(d) and $\tau = 5.0$ for (e), (f). MCP (g) has 4 change-points, which are located at 0.2861, 1.0263, 1.5904, and 2.6647. The GP regression is shown with a solid orange line, and blue shading represents a 95\% uncertainty bound. Illustrated scenarios include (a) $\text{MJD}_{t_{no}}$: An MJD model without drift; (b) $\text{MJD}_{t_{up}}$: MJD model with a globally increasing trend; (c) $\text{MJD}_{t_{inv}}$:  MJD characterized by an increasing trend prior to the shift, followed by a diverging trend post-shift; (d) $\text{MJD}_{t_{down}}$: MJD with an upward trend but a downward shift; (e) $\text{MJD}_{p_{no}}$: MJD with polynomial drift; (f) $\text{MJD}_{p_{up}}$: MJD model with polynomial drift and a globally increasing trend; (g) MCP (K = 4).}
  \label{fig:JDP}
\end{figure*}
%%%%%%%%%%%%%%%%%%%%%%%%%%%%%%%%%%%%%%%%%

\begin{figure*}[t]
% \vspace{.3in}
  \centering
  \begin{subfigure}[b]{0.21\linewidth}
    \includegraphics[width=\linewidth]{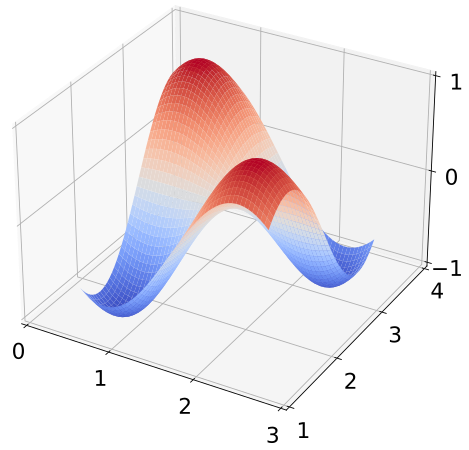}
    \caption{Test function}
  \end{subfigure}
  \begin{subfigure}[b]{0.19\linewidth}
    \includegraphics[width=\linewidth]{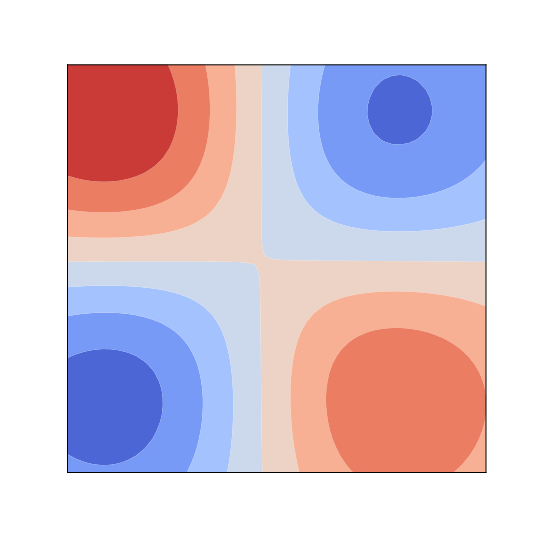}
    \caption{i=0}
  \end{subfigure}
  \begin{subfigure}[b]{0.19\linewidth}
    \includegraphics[width=\linewidth]{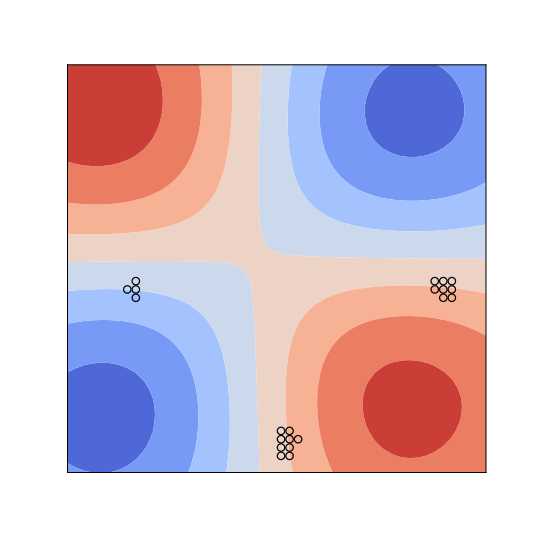}
    \caption{i=20}
  \end{subfigure}
  \begin{subfigure}[b]{0.19\linewidth}
    \includegraphics[width=\linewidth]{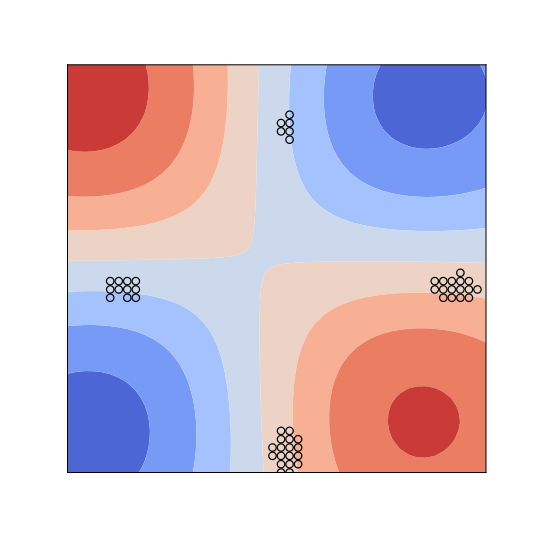}
    \caption{i=50}
  \end{subfigure}
    \begin{subfigure}[b]{0.19\linewidth}
    \includegraphics[width=\linewidth]{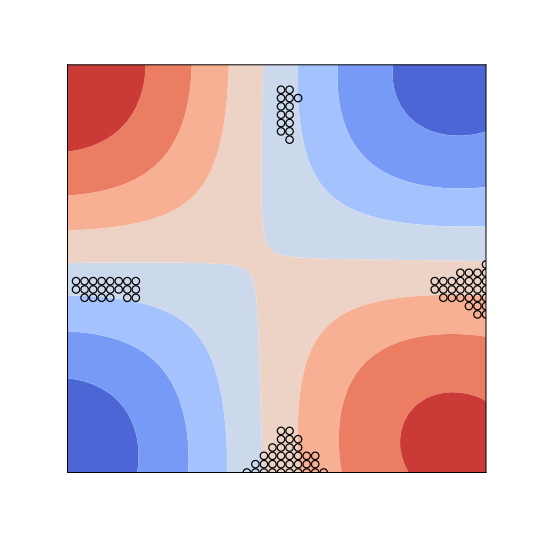}
    \caption{i=99}
  \end{subfigure}
  \begin{subfigure}[b]{0.21\linewidth}
    \includegraphics[width=\linewidth]{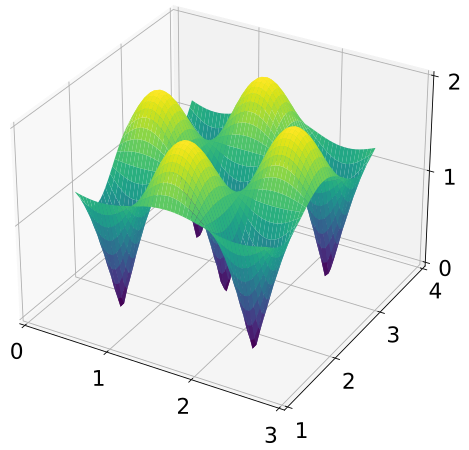}
     \caption{True Derivative}
  \end{subfigure}
  \begin{subfigure}[b]{0.19\linewidth}
    \includegraphics[width=\linewidth]{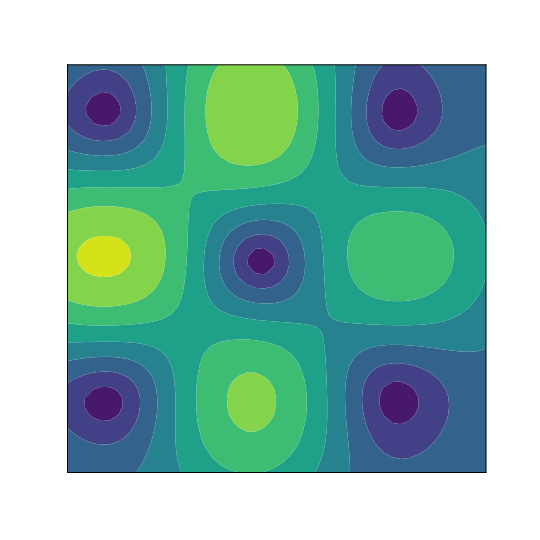}
    \caption{i=0}
  \end{subfigure}
  \begin{subfigure}[b]{0.19\linewidth}
    \includegraphics[width=\linewidth]{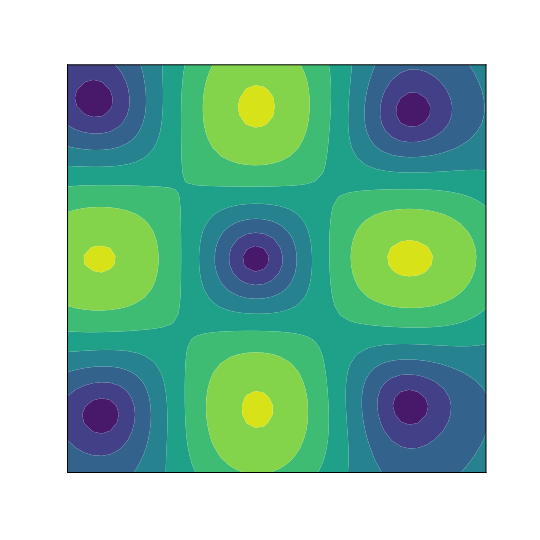}
    \caption{i=20}
  \end{subfigure}
  \begin{subfigure}[b]{0.19\linewidth}
    \includegraphics[width=\linewidth]{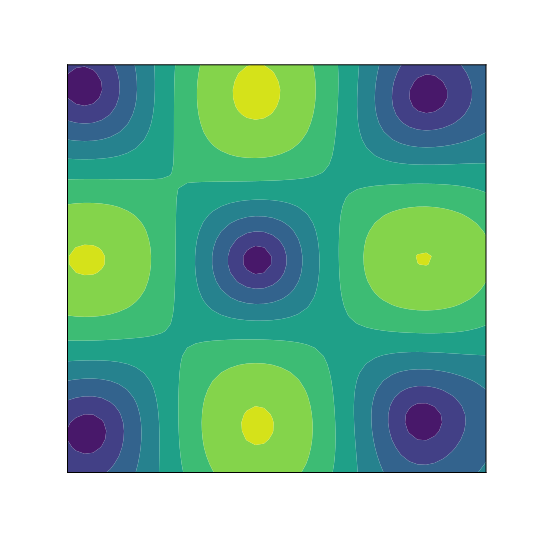}
    \caption{i=50}
  \end{subfigure}
    \begin{subfigure}[b]{0.19\linewidth}
    \includegraphics[width=\linewidth]{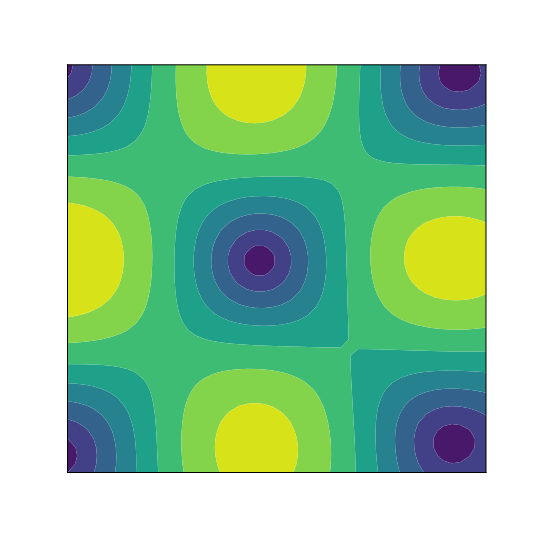}
    \caption{i=99}
  \end{subfigure}
  % \vspace{.3in}
  \caption{Visualization of DACD in 2-dimensional data: The upper row illustrates the multivariate Gaussian distribution with (a) providing a 3D plot of the test function and (b) displaying the predicted mean function post-initial sampling. Subfigures (c)-(e) reveal the predicted mean function at interactions 20, 50, and 99, respectively, with black circles denoting sampled data during AL loops. The lower row depicts the derivative process, with (a) showing the actual derivative distribution and (g)-(j) showing predicted derivatives at specific iterations. Notably, the algorithm effectively chooses points where the abrupt change happened.}
  \label{fig:2D}
\end{figure*}
%%%%%%%%%%%%%%%%%%%%%%%%%%%%%%%%%%%%%%%%%

We examine the CPD accuracy under a sampling cost constraint and utilize both synthetic datasets and real-world datasets to conduct an extensive study of the empirical performance of DACD.

%%%%%%%%%%%%%%%%%%%%%%%%%%%%%%%%%%%%%%%%%%%%%%%%%%%%%%%%%%%%
\subsection{Experiments on Synthetic Datasets}

To evaluate our proposed algorithm, we utilize the Merton Jump-Diffusion (MJD) model \citep{merton1976option}, which was originally for modeling stock price $S_t$ at time $t$. Here, we simply use this flexible model to simulate different underlying process scenarios for evaluating the performance of DACD. The MJD model is governed by a Stochastic Differential Equation (SDE) as follows:
\[
d S_t=\mu S_{t} d t+\sigma S_{t} d W_t+S_{t} d J_t.\]
Here, $\mu$ is the diffusion drift, $\sigma$ represents the diffusion's volatility, $\{W_t\}_{t\geq0}$ is a standard Brownian motion, and $J_t= \sum_{i=1}^{N_t}Y_i$ is a compound Poisson process. The jump sizes $Y_i$ are identically and independently distributed with a distribution of $N(\alpha, \delta^2)$, and the number of jumps $N_t$ follows a Poisson process with jump intensity $\lambda$. The process $S_t$ follows a geometric Brownian motion between jumps. This SDE has an exact solution, which is $S_t=S_0 \exp \left\{\mu t+\sigma W_t-\sigma^2 t / 2+J_t\right\}$. For the nonlinear jump-diffusion model, we adopt:
\[
d S_t=f(S_{t},t) d t+g(S_{t},t) d W_t+h(S_{t},t) d J_t.
\]

We assume the initial observation to be $S_0=1$ and set the time horizon $T=10$. By setting a time step of $dt = 1e-2$, we generate a total of 1000 simulated observations. We subsequently create seven distinct scenarios using the MJD model, each characterized by different patterns and abrupt change behaviors: (1) $\text{MJD}_{t_{no}}$ features no trend; (2) $\text{MJD}_{t_{up}}$ displays an increasing trend; (3) $\text{MJD}_{t_{inv}}$ exhibits diverging trends before and after the change-point; and (4) $\text{MJD}_{t_{down}}$ undergoes a downward shift. Additionally, we examine two polynomial drift variants: (5) $\text{MJD}_{p_{no}}$, which integrates a polynomial drift function, and (6) $\text{MJD}_{p_{up}}$, which combines a polynomial drift with a globally increasing trend. And (7), a MCP case, where $T=4$ and time step $dt = 1e-3$. The specific model parameters employed for creating these simulation scenarios are detailed in Table~\ref{MJD-para}. Note that these underlying processes are unknown and unobservable to an analyst. Assuming the sampling cost is high, we can only take a few observations and use them to infer the underlying process and the possible location of change-points. 

% Notably, the noise level associated with $\text{MJD}_{t_{down}}$ is considerably higher than that of the other MJD time series. This characteristic is expected to significantly influence the range of discontinuities and trend changes detectable by the algorithm.
%%%%%%%%%%%%%%%%%%%%%%%%%%%%%
\begin{table}[h]
\caption{Synthetic functions parameters for different MJD models.}
\centering
\label{MJD-para}
\setlength{\tabcolsep}{3pt}
\begin{tabular}{@{}l*{6}{S[table-format=2.2]}@{}}
\multicolumn{7}{c}{Linear jump-diffusion models}\\
\toprule
  & {$\mu_{pre}$} & {$\mu_{post}$} &{$\sigma$} & {$\lambda$} &{$\alpha$}& {$\delta$}\\ 
\midrule
$\text{MJD}_{t_{no}}$   & {0.00} & {0.00} & 0.10 & 0.18 &0.20 &0.35 \\
$\text{MJD}_{t_{up}}$   & {0.10} & {0.10} & 0.08 & 0.18  &0.30    & 0.35 \\
$\text{MJD}_{t_{inv}}$   & {0.10} & {-0.05} & 0.08 & 0.18 &0.60&0.01\\
$\text{MJD}_{t_{down}}$    & {0.06} & {0.06} & 0.08 & 0.18 &{-0.50} & 0.01\\
MCP  & {0.30} &{-0.40} & 0.30 & 0.90 &{-0.10}&0.70\\ 
\bottomrule
\\
\multicolumn{7}{c}{Nonlinear jump-diffusion models} \\
\toprule
 & \multicolumn{3}{c}{$f(\cdot)$} & \multicolumn{2}{c}{$g(\cdot)$} & {$h(\cdot)$}\\ 
\midrule
$\text{MJD}_{p_{no}}$ & \multicolumn{3}{c}{0.01$S_{t}$(1-$S_{t}$)} &\multicolumn{2}{c}{0.1$\sqrt{S_{t}}$ }&0.60\\
$\text{MJD}_{p_{up}}$ & \multicolumn{3}{c}{0.1$S_{t}$(1-$S_{t}$)+0.041t} &\multicolumn{2}{c}{0.14$\sqrt{S_{t}}$} &0.50\\
\bottomrule
\end{tabular}
\end{table}
%%%%%%%%%%%%%%%%%%%%%%%%%%%%%%%%%%%%%%%%%
We start with an initial sample set $\mathcal{D}$ of eight points. Among these, two are positioned at the boundary, while the remaining six are obtained through random sampling, as depicted in Figure~\ref{fig:JDP}. Following the preliminary setup, we employ GP regression and calculate GP derivatives. From here, we put our attention on the selection of the next sample. This decision hinges on the AF, which serves as our guiding principle. On one hand, exploration presents an opportunity, wherein our choice might gravitate towards the points where the derivative of the covariance is high, highlighting the area with high uncertainty. Conversely, by exploitation, we might select the points with higher mean derivatives, identifying the area with substantial information about abrupt shifts. After exhausting the budget for collecting additional data points, we apply the Filtered Derivative method to determine the change-points, with a sliding window size $A = 100$. We run 20 iterations for single change-point and 100 for MCP. To validate our results, we conducted 100 simulation runs for single change-point and 30 for MCP.

There are few AL methods for CPD in the literature. One of them is the Active Change-point Detection (ACPD) method proposed by \citet{hayashi2019active}. Like ours, this method also sequentially evaluates a limited number of samples to determine a change point. Unlike ours, this method uses traditional parametric time series models, instead of nonparameteric models such as GPs. Thus, DACD will be compared with ACPD in terms of their performance on different test scenarios. We construct an initial observation set with 8 randomly sampled points, then perform an AL loop for 20 iterations (100 for MCP). The estimated change-point is chosen where the change score is maximized.  

%%%%%%%%%%%%%%%%%%%%%%%%%%%%%%%%%%%%%%%%%
\begin{table*}[t]
  \caption{F1 Score Comparison: DACD using various AFs versus baseline ACPD method. Results are based on 100 simulations for single change-point detection and 30 for MCP. The top-performing method for each scenario is emphasized in bold.}
  \label{f1}
  \centering
\begin{tabular}{@{}lccccccc@{}}
\toprule
Method                 & $\text{MJD}_{t_{no}}$ & $\text{MJD}_{t_{up}}$ & $\text{MJD}_{t_{inv}}$ & $\text{MJD}_{t_{down}}$ & $\text{MJD}_{p_{no}}$ & $\text{MJD}_{p_{up}}$ & MCP \\ 
\midrule
DACD ($\text{EI}_{\xi=0.001}$)  & $\boldsymbol{0.83}$             &    0.89        &            0.94   &    $\boldsymbol{0.75}$  & $\boldsymbol{0.87}$  & $\boldsymbol{0.76}$   &0.73 \\
DACD ($\text{EI}_{\xi=0.01}$)   &     0.81         &        0.90   &            0.93   &    0.72 &0.78 &0.75     & 0.69  \\
DACD ($\text{EI}_{\xi=0.1}$)    &   0.77           &      0.81        &          0.87     &      0.72  & 0.68 & 0.62     &  0.68 \\
DACD ($\text{PI}_{\xi=0.075}$)  &   $\boldsymbol{0.83} $    &  $\boldsymbol{0.92}$            &        $\boldsymbol{0.96}$       &    0.70 & 0.79   &   $\boldsymbol{0.76}$  &   $\boldsymbol{0.80}$ \\
DACD ($\text{PI}_{\xi=0.5}$)    &   0.67           &         0.68     &       0.82        &     0.54 & 0.50 &0.44      &  0.77 \\
DACD ($\text{PI}_{\xi=1}$)     &    0.59          &         0.63     &         0.72      &     0.48   & 0.36 & 0.38     &   0.66 \\
DACD ($\text{UCB}_{\lambda=2}$) &    0.79          &       0.88       &          0.94     &      0.68  & 0.70 & 0.72       & 0.70 \\
DACD ($\text{UCB}_{\lambda=4}$) &    0.68         &          0.84    &         0.86      &       0.52  & 0.52 &0.51      & 0.69\\
DACD ($\text{UCB}_{\lambda=8}$) &      0.55        &          0.66    &          0.79     &     0.44   & 0.41 & 0.38     & 0.57  \\
ACPD              &  0.65   &  0.65    & 0.33   &  0.63   &0.34 & 0.26 & 0.11        \\ 
\bottomrule\\
\end{tabular}
\end{table*}
%%%%%%%%%%%%%%%%%%%%%%%%%%%%%%%%%%%%%%%%%

A comprehensive representation of the F1 score calculated from our simulation experiments can be found in Table~\ref{f1}. The DACD algorithm employs the EI, PI, and UCB, with varying hyperparameters. Both PI, with $\xi = 0.075$, and EI, with $\xi = 0.001$, exhibit superior performance across diverse scenarios. This suggests that EI and PI typically serve as more effective AFs for our proposed DACD algorithm. The robust performance of PI and EI can be attributed to its tendency to prioritize exploitation over exploration compared with UCB. However, an increase in hyperparameters reveals a decline in performance, possibly due to a shifting focus from exploitation to exploration. For instance, higher $\xi$ values in PI cause the algorithm to concentrate on areas where predicted improvement is guaranteed, potentially overlooking the exploration of uncertain regions that may contain change-points. Similarly, higher $\lambda$ values in UCB also result in a tilt towards exploration, which may lead to suboptimal performance. Notably, the baseline ACPD method consistently underperforms compared to the DACD methods across all scenarios. Given the assumption that the ACPD method employs the GP with a constant mean, this suggests that the method struggles with convergence to the true change-point.

% The jump-diffusion process, by its nature, is stochastic, signifying that abrupt shifts can occur randomly at any given time with an unknown jump size. In a scenario where the jump size approaches zero, it is the volatility that primarily governs the stochastic process. This situation poses a considerable challenge for the algorithm to accurately detect the true change-point, especially with a limited number of only 28 sample points. However, given an infinite budget for sample collection, the true change-point can indeed be located. The principle behind this is that the abrupt shift is most likely to transpire at the point where the mean derivative is at its highest. In such a context, despite the inherent randomness and volatility, the increase in available data allows for a more accurate estimation of change-points.
%%%%%%%%%%%%%%%%%%%%%%%%%%%%%%%%%%%%%%%%%

\subsection{Experiments on 2-Dimensional Datasets}

\begin{algorithm}[t]
\caption{DACD for 2D Case}
\label{alg:2D}
\renewcommand{\algorithmicrequire}{\textbf{Input:}}
\renewcommand{\algorithmicensure}{\textbf{Output:}}
\begin{algorithmic}[1]
\REQUIRE Set of sample points from AL loops: \( \mathcal{D} \)
\ENSURE Change-points set: $\mathcal{C}$
\FOR{each point \( (\mathbf{x},y) \) in \( \mathcal{D} \)}
    \STATE Determine 10-nearest neighbors \( NN(\mathbf{x}) \)
        \STATE Fit a linear regression \( f \) with points in \( NN(\mathbf{x}) \)
\STATE Obtain the K maximum slope \( S_{\text{max}} \) on \( f \) 
    \STATE Add points to \( \mathcal{C} \)
\ENDFOR
\RETURN\( \mathcal{C} \)
\end{algorithmic}
\end{algorithm}

We consider a 2-dimensional test function (Figure~\ref{fig:2D}) defined on a mesh grid $D=A \times B$, where $A\in[-0.1,3.1]$ and $B \in[0.9, 4.1]$. The grid point set is obtained by dividing the intervals A and B into 50 equal parts. Let 
$\mathcal{X} = \{\mathbf{x} = (x_1,x_2)\in D \,|\, x_1\in[0,3], x_2\in[1,4]\}$, and
\[
    f(x_1, x_2) = \sin(2x_1)\cos(2x_2).
\]
We added an independent Gaussian noise $\mathcal{N}(0, 0.1)$ on each response value. Initially, 100 points are uniformly sampled and their response values are obtained. Then, based on the AFs, AL loops are executed up to iteration 200. The evaluation of the gradient at a sampled point was conducted by identifying its spatial 10-nearest neighbors (KNN) and fitting a linear regression to these neighbors. The locations of the maximum slope of the regression plane are then strategically pinpointed, identifying them as potential change-points, as demonstrated in {\bf{Algorithm}~\ref{alg:2D}}.

% Ground-truths were obtained from the peak values of the true derivative function, with change-points set at 75, representing 3\% of the entire dataset. This process was repeated 50 times, and the average F1 score was calculated, as shown in Figure~\ref{f1}. The results demonstrate that DACD outperforms other methods in terms of performance.

%%%%%%%%%%%%%%%%%%%%%%%%%%%%%%%%%%%%%%%%%%%%%%%%%%%%%%%%%%%%
\subsection{Experiments on Real-World Datasets}

%%%%%%%%%%%%%%%%%%%%%%%%%%%%%%%%%%%%%%%%%
\begin{figure}[t]
\centering
\includegraphics[width=0.99\linewidth]{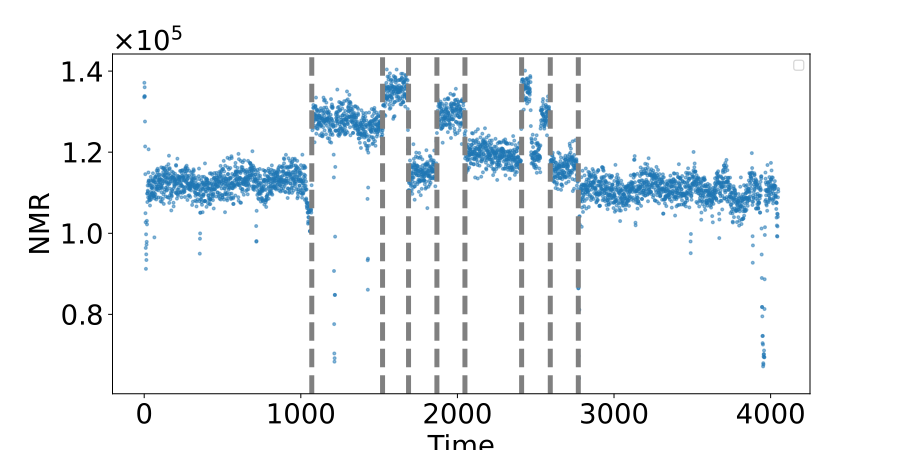}
% \vspace{.3in}
\caption{Well-log Data. }
\label{fig:well_log}
\end{figure}
In this subsection, we focus on a real-world dataset, the well-log data \citep{van2020evaluation, bhatt2022offline, turner2012gaussian}, which has been previously discussed in the literature. This series contains 4050 nuclear magnetic response measurements, as shown in Figure~\ref{fig:well_log}, during well drilling activities. The majority of the observations behave approximately piece-wise constant, but a few data subsets deviate substantially from the mean value. Our revised algorithm can estimate these change-points by identifying the eight most pronounced mean differences within a sliding window. The outcome validates the effectiveness of our method in detecting multiple significant shifts.

%%%%%%%%%%%%%%%%%%%%%%%%%%%%%%%%%%%%%%%%%%%%%%%%%%%%%%%%%%%%
\section{CONCLUSION}
\label{5}
In summary, the DACD method leverages the strengths of the GP and AL to provide an efficient and robust CPD approach, particularly when data acquisition is constrained by time or resources. Through multiple simulations and real-world experiments, we demonstrate that DACD can effectively detect change-points even with a limited number of initial data. This is achieved by integrating GP derivatives with a balanced exploration-exploitation strategy into the AF, making it a valuable tool for a wide range of applications.

%%%%%%%%%%%%%%%%%%%%%%%%%%%%%%%%%%%%%%%%%%%%%%%%%%%%%%%%%%%%

% \subsubsection*{Acknowledgements}
% All acknowledgments go at the end of the paper, including thanks to reviewers who gave useful comments, to colleagues who contributed to the ideas, and to funding agencies and corporate sponsors that provided financial support. 
% To preserve the anonymity, please include acknowledgments \emph{only} in the camera-ready papers.

\bibliography{refs}

\begin{thebibliography}{}

\bibitem[Ahmadzadeh, 2018]{ahmadzadeh2018change}
Ahmadzadeh, F. (2018).
\newblock Change point detection with multivariate control charts by artificial neural network.
\newblock {\em The International Journal of Advanced Manufacturing Technology}, 97:3179--3190.

\bibitem[Aminikhanghahi and Cook, 2017]{aminikhanghahi2017survey}
Aminikhanghahi, S. and Cook, D.~J. (2017).
\newblock A survey of methods for time series change point detection.
\newblock {\em Knowledge and information systems}, 51(2):339--367.

\bibitem[Appel and Brandt, 1983]{appel1983adaptive}
Appel, U. and Brandt, A.~V. (1983).
\newblock Adaptive sequential segmentation of piecewise stationary time series.
\newblock {\em Information sciences}, 29(1):27--56.

\bibitem[Basseville et~al., 1993]{basseville1993detection}
Basseville, M., Nikiforov, I.~V., et~al. (1993).
\newblock {\em Detection of abrupt changes: theory and application}, volume 104.
\newblock prentice Hall Englewood Cliffs.

\bibitem[Bertrand et~al., 2011]{bertrand2011off}
Bertrand, P.~R., Fhima, M., and Guillin, A. (2011).
\newblock Off-line detection of multiple change points by the filtered derivative with p-value method.
\newblock {\em Sequential Analysis}, 30(2):172--207.

\bibitem[Bhatt et~al., 2022]{bhatt2022offline}
Bhatt, S., Fang, G., and Li, P. (2022).
\newblock Offline change detection under contamination.
\newblock In {\em Uncertainty in Artificial Intelligence}, pages 191--201. PMLR.

\bibitem[De~Brabandere et~al., 2022]{de2022semi}
De~Brabandere, A., Cao, Z., De~Vos, M., Bertrand, A., and Davis, J. (2022).
\newblock Semi-supervised change point detection using active learning.
\newblock In {\em Discovery Science: 25th International Conference, DS 2022, Montpellier, France, October 10--12, 2022, Proceedings}, pages 74--88. Springer.

\bibitem[Gijbels and Goderniaux, 2005]{gijbels2005data}
Gijbels, I. and Goderniaux, A.-C. (2005).
\newblock Data-driven discontinuity detection in derivatives of a regression function.
\newblock {\em Communications in Statistics-Theory and Methods}, 33(4):851--871.

\bibitem[Grzesiek et~al., 2021]{grzesiek2021method}
Grzesiek, A., Zimroz, R., {\'S}liwi{\'n}ski, P., Gomolla, N., and Wy{\l}oma{\'n}ska, A. (2021).
\newblock A method for structure breaking point detection in engine oil pressure data.
\newblock {\em Energies}, 14(17):5496.

\bibitem[Hayashi et~al., 2019]{hayashi2019active}
Hayashi, S., Kawahara, Y., and Kashima, H. (2019).
\newblock Active change-point detection.
\newblock In {\em Asian Conference on Machine Learning}, pages 1017--1032. PMLR.

\bibitem[Inatsu et~al., 2020]{inatsu2020active}
Inatsu, Y., Sugita, D., Toyoura, K., and Takeuchi, I. (2020).
\newblock Active learning for enumerating local minima based on gaussian process derivatives.
\newblock {\em Neural Computation}, 32(10):2032--2068.

\bibitem[Johnson et~al., 2020]{johnson2020kernel}
Johnson, J.~E., Laparra, V., P{\'e}rez-Suay, A., Mahecha, M.~D., and Camps-Valls, G. (2020).
\newblock Kernel methods and their derivatives: Concept and perspectives for the earth system sciences.
\newblock {\em Plos one}, 15(10):e0235885.

\bibitem[Kushner, 1964]{kushner1964new}
Kushner, H.~J. (1964).
\newblock A new method of locating the maximum point of an arbitrary multipeak curve in the presence of noise.

\bibitem[McHutchon et~al., 2015]{mchutchon2015nonlinear}
McHutchon, A.~J. et~al. (2015).
\newblock {\em Nonlinear modelling and control using Gaussian processes}.
\newblock PhD thesis, Citeseer.

\bibitem[Merton, 1976]{merton1976option}
Merton, R.~C. (1976).
\newblock Option pricing when underlying stock returns are discontinuous.
\newblock {\em Journal of financial economics}, 3(1-2):125--144.

\bibitem[Mockus, 1998]{mockus1998application}
Mockus, J. (1998).
\newblock The application of bayesian methods for seeking the extremum.
\newblock {\em Towards global optimization}, 2:117.

\bibitem[Myers et~al., 2016]{myers2016response}
Myers, R.~H., Montgomery, D.~C., and Anderson-Cook, C.~M. (2016).
\newblock {\em Response surface methodology: process and product optimization using designed experiments}.
\newblock John Wiley \& Sons.

\bibitem[Nguyen et~al., 2017]{nguyen2017regret}
Nguyen, V., Gupta, S., Rana, S., Li, C., and Venkatesh, S. (2017).
\newblock Regret for expected improvement over the best-observed value and stopping condition.
\newblock In {\em Asian conference on machine learning}, pages 279--294. PMLR.

\bibitem[Oelsmann et~al., 2022]{oelsmann2022bayesian}
Oelsmann, J., Passaro, M., S{\'a}nchez, L., Dettmering, D., Schwatke, C., and Seitz, F. (2022).
\newblock Bayesian modelling of piecewise trends and discontinuities to improve the estimation of coastal vertical land motion: Discotimes: a method to detect change points in gnss, satellite altimetry, tide gauge and other geophysical time series.
\newblock {\em Journal of Geodesy}, 96(9):62.

\bibitem[Padidar et~al., 2021]{padidar2021scaling}
Padidar, M., Zhu, X., Huang, L., Gardner, J., and Bindel, D. (2021).
\newblock Scaling gaussian processes with derivative information using variational inference.
\newblock {\em Advances in Neural Information Processing Systems}, 34:6442--6453.

\bibitem[Pan and Rigdon, 2012]{pan2012bayesian}
Pan, R. and Rigdon, S.~E. (2012).
\newblock A bayesian approach to change point estimation in multivariate spc.
\newblock {\em Journal of Quality Technology}, 44(3):231--248.

\bibitem[Pepelyshev and Polunchenko, 2015]{pepelyshev2015real}
Pepelyshev, A. and Polunchenko, A.~S. (2015).
\newblock Real-time financial surveillance via quickest change-point detection methods.
\newblock {\em arXiv preprint arXiv:1509.01570}.

\bibitem[Rasmussen et~al., 2006]{rasmussen2006gaussian}
Rasmussen, C.~E., Williams, C.~K., et~al. (2006).
\newblock {\em Gaussian processes for machine learning}, volume~1.
\newblock Springer.

\bibitem[Ropkins and Tate, 2021]{ropkins2021early}
Ropkins, K. and Tate, J.~E. (2021).
\newblock Early observations on the impact of the covid-19 lockdown on air quality trends across the uk.
\newblock {\em Science of the Total Environment}, 754:142374.

\bibitem[Schiepek et~al., 2020]{schiepek2020convergent}
Schiepek, G., Sch{\"o}ller, H., de~Felice, G., Steffensen, S.~V., Bloch, M.~S., Fartacek, C., Aichhorn, W., and Viol, K. (2020).
\newblock Convergent validation of methods for the identification of psychotherapeutic phase transitions in time series of empirical and model systems.
\newblock {\em Frontiers in Psychology}, 11:1970.

\bibitem[Settles, 2009]{settles2009active}
Settles, B. (2009).
\newblock Active learning literature survey.

\bibitem[Shahriari et~al., 2015]{shahriari2015taking}
Shahriari, B., Swersky, K., Wang, Z., Adams, R.~P., and De~Freitas, N. (2015).
\newblock Taking the human out of the loop: A review of bayesian optimization.
\newblock {\em Proceedings of the IEEE}, 104(1):148--175.

\bibitem[Shen et~al., 2015]{shen2015impact}
Shen, Z., Hou, X., Li, W., Aini, G., Chen, L., and Gong, Y. (2015).
\newblock Impact of landscape pattern at multiple spatial scales on water quality: A case study in a typical urbanised watershed in china.
\newblock {\em Ecological Indicators}, 48:417--427.

\bibitem[Solak et~al., 2002]{solak2002derivative}
Solak, E., Murray-Smith, R., Leithead, W., Leith, D., and Rasmussen, C. (2002).
\newblock Derivative observations in gaussian process models of dynamic systems.
\newblock {\em Advances in neural information processing systems}, 15.

\bibitem[Srinivas et~al., 2009]{srinivas2009gaussian}
Srinivas, N., Krause, A., Kakade, S.~M., and Seeger, M. (2009).
\newblock Gaussian process optimization in the bandit setting: No regret and experimental design.
\newblock {\em arXiv preprint arXiv:0912.3995}.

\bibitem[Toodesh et~al., 2021]{toodesh2021prediction}
Toodesh, R., Verhagen, S., and Dagla, A. (2021).
\newblock Prediction of changes in seafloor depths based on time series of bathymetry observations: Dutch north sea case.
\newblock {\em Journal of Marine Science and Engineering}, 9(9):931.

\bibitem[Truong et~al., 2020]{truong2020selective}
Truong, C., Oudre, L., and Vayatis, N. (2020).
\newblock Selective review of offline change point detection methods.
\newblock {\em Signal Processing}, 167:107299.

\bibitem[Turner, 2012]{turner2012gaussian}
Turner, R.~D. (2012).
\newblock {\em Gaussian processes for state space models and change point detection}.
\newblock PhD thesis, University of Cambridge.

\bibitem[Van~den Burg and Williams, 2020]{van2020evaluation}
Van~den Burg, G.~J. and Williams, C.~K. (2020).
\newblock An evaluation of change point detection algorithms.
\newblock {\em arXiv preprint arXiv:2003.06222}.

\bibitem[Verbesselt et~al., 2010]{verbesselt2010detecting}
Verbesselt, J., Hyndman, R., Newnham, G., and Culvenor, D. (2010).
\newblock Detecting trend and seasonal changes in satellite image time series.
\newblock {\em Remote sensing of Environment}, 114(1):106--115.

\bibitem[Vezhnevets et~al., 2012]{vezhnevets2012active}
Vezhnevets, A., Buhmann, J.~M., and Ferrari, V. (2012).
\newblock Active learning for semantic segmentation with expected change.
\newblock In {\em 2012 IEEE conference on computer vision and pattern recognition}, pages 3162--3169. IEEE.

\bibitem[Wu et~al., 2017]{wu2017bayesian}
Wu, J., Poloczek, M., Wilson, A.~G., and Frazier, P. (2017).
\newblock Bayesian optimization with gradients.
\newblock {\em Advances in neural information processing systems}, 30.

\bibitem[Ye et~al., 2023]{ye2023online}
Ye, H., Xian, X., Cheng, J.-R.~C., Hable, B., Shannon, R.~W., Elyaderani, M.~K., and Liu, K. (2023).
\newblock Online nonparametric monitoring of heterogeneous data streams with partial observations based on thompson sampling.
\newblock {\em IISE Transactions}, 55(4):392--404.

\bibitem[Yue et~al., 2020]{yue2020active}
Yue, X., Wen, Y., Hunt, J.~H., and Shi, J. (2020).
\newblock Active learning for gaussian process considering uncertainties with application to shape control of composite fuselage.
\newblock {\em IEEE Transactions on Automation Science and Engineering}, 18(1):36--46.

\end{thebibliography}

\end{document}